%% file: main.tex
\documentclass[letterpaper]{article} 
\usepackage[]{aaai25}  
\usepackage{times}  
\usepackage{helvet}  
\usepackage{courier}  
\usepackage[hyphens]{url}  
\usepackage{graphicx} 
\usepackage{natbib}  
\usepackage{caption} 
\usepackage{algorithm}
\usepackage{algpseudocode}

\usepackage{newfloat}
\usepackage{listings}
\DeclareCaptionStyle{ruled}{labelfont=normalfont,labelsep=colon,strut=off} 
\floatstyle{ruled}
\newfloat{listing}{tb}{lst}{}
\floatname{listing}{Listing}
\input{macro}
\usepackage{enumitem}

\title{Show, Don't Tell: Learning Reward Machines from Demonstrations for Reinforcement Learning-Based Cardiac Pacemaker Synthesis}
\author{
    John Komp\textsuperscript{\rm 1}, Dananjay Srinivas\textsuperscript{\rm 1}, Maria Pacheco\textsuperscript{\rm 1}, Ashutosh Trivedi\textsuperscript{\rm 1}}

\affiliations{
    \textsuperscript{\rm 1}University of Colorado Boulder \\

    \{john.komp, Dananjay.Srinivas, maria.pacheco, ashutosh.trivedi\}@colorado.edu
}

\begin{document}

\maketitle
\thispagestyle{plain}
\pagestyle{plain}

\begin{abstract}
    \input{abstract}

\end{abstract}

\section{Introduction}
\input{introduction}

\section{Basics of Pacing Therapy}
\label{sec:definition}
\input{preliminaries}

\section{Learning Reward Machines}
\label{sec:rm}
\input{learning_rewards}

\section{Case Study}
\label{sec:casestudy}
\input{experiments}

\section{Related Work}
\label{sec:related}
\input{related.tex}

\section{Conclusion}
\label{sec:conclusion}
\input{conclusions}

\bibliography{papers}

\appendix

\section{Reproducibility Checklist}

\begin{itemize}[left=0pt]

    \item \textbf{This paper:}
    \begin{itemize}
        \item Includes a conceptual outline and/or pseudocode description of AI methods introduced (yes)
        \item Clearly delineates statements that are opinions, hypothesis, and speculation from objective facts and results (yes)
        \item Provides well-marked pedagogical references for less-familiar readers to gain the background necessary to replicate the paper (yes)
    \end{itemize}

    \item \textbf{Does this paper make theoretical contributions?} (no)

    \item \textbf{Does this paper rely on one or more datasets?} (yes)
    \begin{itemize}
        \item If yes, please complete the list below.
        \begin{itemize}
            \item A motivation is given for why the experiments are conducted on the selected datasets (yes)
            \item All novel datasets introduced in this paper are included in a data appendix. (partial)
            \item All novel datasets introduced in this paper will be made publicly available upon publication of the paper with a license that allows free usage for research purposes. (yes)
            \item All datasets drawn from the existing literature (potentially including authors’ own previously published work) are accompanied by appropriate citations. (yes)
            \item All datasets drawn from the existing literature (potentially including authors’ own previously published work) are publicly available. (yes)
            \item All datasets that are not publicly available are described in detail, with an explanation of why publicly available alternatives are not scientifically satisfying. (yes)
        \end{itemize}
    \end{itemize}

    \item \textbf{Does this paper include computational experiments?} (yes)
    \begin{itemize}
        \item If yes, please complete the list below.
        \begin{itemize}
            \item Any code required for pre-processing data is included in the appendix. (yes)
            \item All source code required for conducting and analyzing the experiments is included in a code appendix. (partial)
            \item All source code required for conducting and analyzing the experiments will be made publicly available upon publication of the paper with a license that allows free usage for research purposes. (yes)
            \item All source code implementing new methods has comments detailing the implementation, with references to the paper where each step comes from (partial)
            \item If an algorithm depends on randomness, the method used for setting seeds is described in a way sufficient to allow replication of results. (yes)
            \item This paper specifies the computing infrastructure used for running experiments (hardware and software), including GPU/CPU models; amount of memory; operating system; names and versions of relevant software libraries and frameworks. (partial)
            \item This paper formally describes evaluation metrics used and explains the motivation for choosing these metrics. (yes)
            \item This paper states the number of algorithm runs used to compute each reported result. (yes)
            \item Analysis of experiments goes beyond single-dimensional summaries of performance (e.g., average; median) to include measures of variation, confidence, or other distributional information. (yes)
            \item The significance of any improvement or decrease in performance is judged using appropriate statistical tests (e.g., Wilcoxon signed-rank). (no)
            \item This paper lists all final (hyper-)parameters used for each model/algorithm in the paper’s experiments. (partial)
            \item This paper states the number and range of values tried per (hyper-)parameter during the development of the paper, along with the criterion used for selecting the final parameter setting. (yes)
        \end{itemize}
    \end{itemize}

\end{itemize}

 \input{appendix.tex}

\end{document}

%% file: macro.tex
\usepackage[dvipsnames]{xcolor} 

\usepackage{amsmath}
\usepackage{amssymb}

\usepackage{adjustbox}
\usepackage{multirow}
\usepackage{booktabs}
\usepackage{subcaption}
\usepackage{tikz}
\usetikzlibrary{arrows,shapes}

\newcommand{\Var}{\mathit{Var}}

\newcommand{\Nat}{\mathbb N}

\newcommand{\true}{\mathsf{true}}
\newcommand{\false}{\mathsf{false}}

\newcommand{\llimit}{\texttt{lrl}}
\newcommand{\ulimit}{\texttt{url}}

\newcommand{\seq}[1]{\langle #1 \rangle}
\newcommand{\set}[1]{\left\{ #1 \right\}}
\newcommand{\chop}{^{\frown}}
\newcommand{\mop}{\bowtie}
\newcommand{\pointP}[1]{\lceil #1 \rceil^{\bullet}{}}

\newcommand{\rangeP}[1]{\lceil #1 \rceil{}}
\newcommand{\ftrigInt}[3]{#1\hookrightarrow{}_{#2}#3}

\def\rmdef{\stackrel{\mbox{\rm {\tiny def}}}{=}} 

%% file: abstract.tex
An (artificial cardiac) pacemaker is an implantable electronic device that sends electrical impulses to the heart to regulate the heartbeat.  
As the number of pacemaker users continues to rise, so does the demand for features with additional sensors, adaptability, and improved battery performance. Reinforcement learning (RL) has recently been proposed as a performant algorithm for creative design space exploration, adaptation, and statistical verification of cardiac pacemakers. 
The design of correct reward functions, expressed as a reward machine, is a key programming activity in this process.  

In 2007, Boston Scientific published a detailed description of their pacemaker specifications. 
This document has since formed the basis for several formal characterizations of pacemaker specifications using real-time automata and logic. 
However, because these translations are done manually, they are challenging to verify. 
Moreover, capturing requirements in automata or logic is notoriously difficult.
We posit that it is significantly easier for domain experts, such as electrophysiologists, to observe and identify abnormalities in electrocardiograms that correspond to patient-pacemaker interactions. 
Therefore, we explore the possibility of learning correctness specifications from such labeled demonstrations in the form of a reward machine and training an RL agent to synthesize a cardiac pacemaker based on the resulting reward machine.

We leverage advances in machine learning to extract signals from labeled demonstrations as reward machines using recurrent neural networks and transformer architectures. These reward machines are then used to design a simple pacemaker with RL. Finally, we validate the resulting pacemaker using properties extracted from the Boston Scientific document.

%% file: introduction.tex
The human heart is arguably the most important real-time system. 
When functioning correctly, the heart's natural electrical system sends signals through its chambers to regulate the heartbeat. 
However, due to factors such as heart muscle damage or congenital mutations, this signaling can become irregular. 
In such situations, artificial cardiac pacemakers (henceforth referred to as pacemakers) are routinely prescribed to regulate the heart rhythm.
\emph{This paper presents a novel approach to designing cardiac pacemakers using reinforcement learning and expert demonstrations.}

\paragraph{Advances in Pacemaker Design.} A pacemaker is a tiny, battery-powered device implanted below the collarbone that connects to the heart via sensor and actuator leads or affixed directly into the atrial or ventricular wall. 
Its main job is to supply missing or delayed electrical signals with precise timing, appropriately reacting to the body's resting or active states. 
As the number of patients using pacemakers increases, so do consumer expectations for the ``invisibility'' of its design. 
This demands that manufacturers extend capabilities
~\cite{khan2022innovations}
by integrating more sensors and complex, adaptive pacing logic.
Recently, reinforcement learning (RL) has been proposed for designing cardiac pacemakers and other implantable medical devices due to its capabilities in creative design exploration, automated testing, and adaptive solutions~\cite{dole2023correct,datta2021artificial,
rom2011optimal}.

\paragraph{Design and Verification of Pacemakers.} Traditionally, pacemaker design has been a manual process where engineers rely on their expertise to create devices that meet clinical needs. However, this approach is increasingly supplemented by model-based design, which uses computational models to simulate, test, and optimize device behavior before physical prototypes are built. This method improves correctness and reduces development time and costs.
Pacemakers must comply with stringent FDA medical device regulations that ensure safety, efficacy, and reliability~\cite{FDA2020cfr,FDA2002principles,FDA2023Expectations}. These requirements mandate rigorous testing and validation processes, including preclinical trials, clinical trials, and post-market surveillance. The FDA has recently begun issuing guidance for use of artificial intelligence, including RL, in medical devices \cite{FDA2023ai}.
In 2007, Boston Scientific published a comprehensive description of their pacemaker system specifications~\cite{paceSpec2009}. This document has been instrumental in the formal characterization of pacemaker specifications using automata and logic.

\begin{figure*}[t]
    \centering
    \includegraphics[width=13.0cm,height=6.5cm,keepaspectratio]{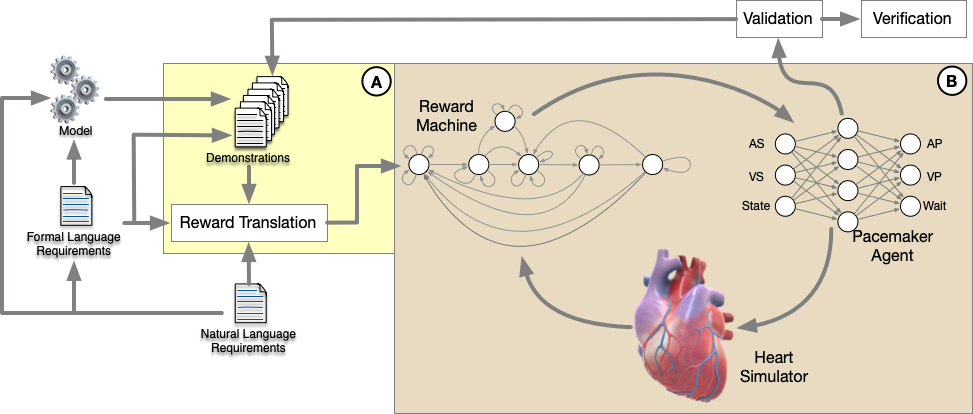}
    \caption{Learning Process from requirements and exemplars to trained pacemaker agent}
    \label{fig:process}
\end{figure*}

\paragraph{RL-based Synthesis.} RL has emerged as a powerful tool for designing adaptive pacemakers due to its capabilities in creative exploration, adaptive solutions, and verification. 
Traditional pacemakers operate based on pre-defined rules and fixed programming, which may not adapt well to the dynamic physiological changes in patients. 
RL, however, enables the development of adaptive pacemakers that can learn from the patient’s real-time physiological data and adjust their behavior accordingly.
RL allows for the exploration of a vast design space, enabling the discovery of innovative solutions that might not be apparent through conventional design methods. 
By repeated sampling and optimizing, RL can uncover novel pacing strategies that improve patient outcomes and device performance. 
Moreover, RL can automate the testing and verification process by simulating a wide range of physiological conditions and responses. 
This helps identify potential issues and ensures that the pacemaker meets the stringent requirements for clinical use.

\paragraph{Reward Machines.} The design of the reward machine is the most critical programming challenge in any RL-based design process. 
Expressing the learning objectives as scalar rewards is primarily a manual and error-prone task. 
While it is possible to automatically compile formal specifications into reward signals~\cite{dole2023correct}, writing formal specifications is inherently difficult. 
Timed automata can be used to capture these specifications~\cite{jiang2012modeling}, but the process remains manual and challenging to verify.
To address these challenges, we leverage advances in machine learning and formal methods to streamline the design process. 
By combining expert demonstrations with techniques for extracting specifications, we aim to create an efficient and reliable framework for RL-based pacemaker design.

\paragraph{Availability of Labelled Traces.} It is considerably easier to label pacemaker-heart closed-loop traces, which are readily available from electrophysiologists (EPs), online repositories~\cite{bhrs24}, previous versions of pacemakers, and digital twins. 
If we can enable the use of demonstrations in RL-based pacemaker design, it will allow us to directly leverage the insights and experience of EPs in designing the next generation of pacemakers. 
This project aims to achieve this goal by integrating labeled traces into the RL framework, thereby enhancing the adaptability, precision, and effectiveness of pacemaker devices. 
By utilizing these expert-labeled demonstrations, we can develop more sophisticated reward machines and improve the overall design process, ensuring that pacemakers better meet the needs of patients.

\vspace{0.4em}\noindent\textbf{Advances in Sequence Learning.} To create reward machines from labelled traces, we build a classifier to predict whether the last pacemaker action in a sequence actions corresponds to acceptable or erroneous behavior. 
This allows us to take advantage of the availability of labelled demonstrations and learn to provide signal to the RL agent at any step during its execution. 
We build on recent advances in sequence learning and implement two modern deep learning classifiers, a long short-term memory recurrent neural network (LSTM)~\cite{lstm} and a transformer neural network~\cite{transformers}. Neural sequence architectures have shown to be successful at modeling heart-related signal, such as electrocardiogram data~\cite{9190034, s22093446}, as well as traces of various real-time systems such brain-machine interfaces~\cite{10.1145/3495006, 9948637}, mobile networks~\cite{8824981}, and various IoT devices~\cite{9838882, MOHAMMED2024103793}. In this paper, we set out to explore whether neural sequence architectures are effective at modeling pacemaker execution.


\paragraph{RL-driven Pacemaker Synthesis.}
Figure \ref{fig:process} illustrates the process of training an RL agent as a bradycardia pacemaker. 
A key element of this training is the creation of a reward machine that guides the training. 
There are multiple ways to create such a machine: through human translation of a requirements specification, modeling of formal requirements, or using traces from an existing model or representative black box. 
Here, we present a method based on the latter approach, where a pacemaker is trained directly from example traces. 
The process has two parts: creating the reward machine, highlighted as Area A in the figure, and training the RL agent using the reward machine, shown as Area B.
The goal of this paper is to assess the practicality of training an RL agent to operate as a complex real-time device directly from operational examples. With that in mind, we pose the following research questions:
\begin{enumerate}[start=1,label={\bfseries RQ\arabic*},leftmargin={3em}]
\item Can sequence learning recognize the proper operation of a real-time device, as a (neural) reward machine, from a sparse dataset representing good and bad operation?
\item Can an RL agent be trained to correctly pace a heart, using the aforementioned reward machine?
\end{enumerate}

\paragraph{Contributions.} The key contributions are listed below.
\begin{enumerate}
    \item We contribute an annotated dataset of 11,000 simulated pacemaker traces that spans correct and incorrect executions across five different arrhythmia types, as well as healthy heart behavior. 
    \item We show that neural sequence learning algorithms are effective at learning to predict the correctness of the next action in a trace, with varying amounts of prior context.
    \item We show that RL agents can learn a complex, safety-critical, real-time controller from only good and bad traces of process execution.
\end{enumerate}
While our focus is on the cardiac pacemaker case study, the techniques presented may be applicable to other settings. These techniques exploit neural sequence architectures (LSTMs and transformers) with deep RL to design complex machines from expert demonstrations.


%% file: preliminaries.tex
\paragraph{Heart Disease.}
The human heart beat is produced by synchronized contractions of the atria and ventricles through a chemically activated circuit in the myocardium tissue. Each beat begins with the sinoatrial (SA) node, located in the upper left portion of the right atrium, initiating a depolarization of the myocardium propagating through the atria causing a contraction. When the depolarization reaches the atrioventricular (AV) node located between the right atrium and ventricle, it is delayed (150-250ms typical for 60 bpm rate) before continuing through the ventricles causing their contraction and completion of the cycle. After each chamber's contraction a repolarization period returns the myocardium to its resting state. 
\citet{kay2017stimulation} provide a rigorous description of the mycardial tissue contraction process.

Heart disease disrupts the synchrony of the atrial and venctricular contractions. 
Bradycardia describes a category of a cardiac disease where the heart does not provide sufficient perfusion support by beating too slowly. Typical symptoms include missing heart beats, irregular heart rates, and lack of rate increase under stress or exercise. A bradycardia patient may syncopate during normal daily activities.


\input{pacemaker_trace}

\paragraph{Bradycardia Pacing Therapy.}
When a patient presents with bradycardia, pacing therapy is the most common treatment to restore healthy cardiac function. A dual chamber pacemaker analyzes the electrical activity in the right atrium and ventricle to detect the intrinsic beating of the heart. When an expected intrinsic beat does not occur, the pacemaker provides an electrical pulse in the appropriate chamber to initiate the missing contraction.

Pacing therapy is defined based on a collection of intervals used to predict what part of the depolarization/repolarization cycle each chamber is currently in. The pacing cycle for dual chamber, bradycardia pacemakers is measured from atrial event, an intrinsic beat or a pacing pulse, to the next atrial event. This period or A-A interval is the \emph{lower rate interval} (LRI), the longest time the pacemaker will wait before pacing the atrium. 
The LRI is divided into two parts, the shorter time from the atrial to ventricular contraction, the \emph{AV interval}, and the larger time from the ventricular to next atrial contraction, the \emph{VA interval}. 
At the end of these intervals, if no intrinsic beat has occurred, the pacemaker will provide an electric shock to the appropriate chamber stimulating the missing contraction. 
A detected intrinsic beat would terminate the interval and start the next (e.g. in the AV interval, an intrinsic ventricular beat would terminate the AV interval and start the VA interval).

There are two subperiods starting at the beginning of each interval for each chamber, \emph{blanking} and \emph{refractory}. 
The pacing timing trace shown in Figure \ref{fig_applicationTimelines} illustrates these periods. 
The blanking period is a span of time where no intrinsic sensing is occurring. For many pacemakers the preceding intrinsic event or pacing pulse may saturate the pacemaker input sensing amplifier and the blanking period allows for recovery and avoid false sensing. 
The refractory period represents the repolarization period of the chamber. Any intrinsic events detected during this period are considered valid but too early to be hemodynamically effective, an electrically reentrant path, or far-field noise. 
Such events are ignored but cause the respective  refractory period timer to restart.


\paragraph{Pacemaker Requirements.}\label{pacemaker_requirements}
The minimum requirements for a bradycardia pacemaker are relatively simple. The device should always pace at the set lower rate when no valid intrinsic beat is detected and it should never pace faster than the set upper rate. As an example of simplicity, the very first battery powered pacemaker used a metronome integrated circuit~\cite{bakken1999} to time the interval between beats. The specification~\cite{paceSpec2009} provides the minimum requirements for any dual chamber pacemaker. Modern pacemakers extend these capabilities to address less common cardiac arrhythmias and tailor treatment to each individual patient.

Pacemaker timing is cyclic, based solely on the LRI. There are no open ended intervals. This allows us to use the rich temporal syntax of Duration Calculus (DC)~\cite{DBLP:conf/stacs/ChaochenHS93} to specify the real-time pacemaker behavior. 
While the general DC is undecidable, taking advantage of the cyclic nature of pacing we can constrain our usage to a time-bound DC fragment, \emph{discrete duration calculus} (DDC)~\cite{dole2023correct} to assure decidability without losing expressiveness.

%% file: pacemaker_trace.tex
\begin{figure*}[t]
  \centering
\adjustbox{max width=13.0cm}{
\begin{tikzpicture}[xscale=0.5,yscale=2.0]
\draw (4,3) -- (4,3+0.2);
\draw (4,3) +(0,0.35) node [text width=1.5cm,font=\sffamily,align=center, text=teal] {AS};
\draw (13,3) -- (13,3+0.2);
\draw (13,3) +(0,0.35) node [text width=1.5cm,font=\sffamily,align=center, text=cyan] {VP};
\draw (33,3) -- (33,3+0.2);
\draw (33,3) +(0,0.35) node [text width=1.5cm,font=\sffamily,align=center, text=teal] {AS};
\foreach \style/\xa/\ya/\xb/\yb in {lightgray/4/3/6/2.8875,lightgray/4/3/6/3.1125,none/-0.25/3/13/3.2,lightgray/13/3/15/3.1125,lightgray/13/3/15/2.8875,none/13/3/17/2.8,none/13/3/18/3.2,lightgray/33/3/35/2.8875,lightgray/33/3/35/3.1125}
\draw[black, fill=\style, thick] (\xa,\ya) rectangle (\xb,\yb);
\draw (4,2) -- (4,2+0.2);
\draw (4,2) +(0,0.35) node [text width=1.5cm,font=\sffamily,align=center, text=teal] {VS};
\draw (25,2) -- (25,2+0.2);
\draw (25,2) +(0,0.35) node [text width=1.5cm,font=\sffamily,align=center, text=cyan] {AP};
\draw (37,2) -- (37,2+0.2);
\draw (37,2) +(0,0.35) node [text width=1.5cm,font=\sffamily,align=center, text=teal] {AS};
\foreach \style/\xa/\ya/\xb/\yb in {none/33/3/39.25/3.2,none/-0.25/2/4/2.2,lightgray/4/2/6/1.8875,lightgray/4/2/6/2.1125,none/4/2/8/1.8,none/4/2/9/2.2,lightgray/25/2/27/2.1125,lightgray/25/2/27/1.8875}
\draw[black, fill=\style, thick] (\xa,\ya) rectangle (\xb,\yb);
\draw (1,1) -- (1,1+0.2);
\draw (1,1) +(0,0.35) node [text width=1.5cm,font=\sffamily,align=center, text=cyan] {VP};
\draw (9,1) -- (9,1+0.2);
\draw (9,1) +(0,0.35) node [text width=1.5cm,font=\sffamily,align=center, text=teal] {AS};
\draw (24,1) -- (24,1+0.2);
\draw (24,1) +(0,0.35) node [text width=1.5cm,font=\sffamily,align=center, text=teal] {VS};
\draw (33,1) -- (33,1+0.2);
\draw (33,1) +(0,0.35) node [text width=1.5cm,font=\sffamily,align=center, text=teal] {AS};
\foreach \style/\xa/\ya/\xb/\yb in {lightgray/37/2/39/1.8875,lightgray/37/2/39/2.1125,none/25/2/39.25/2.2,none/-0.25/1/1/1.2,lightgray/1/1/3/1.1125,lightgray/1/1/3/0.8875,none/1/1/5/0.8,none/1/1/6/1.2,lightgray/9/1/11/0.8875,lightgray/9/1/11/1.1125,none/9/1/24/1.2,lightgray/24/1/26/0.8875,lightgray/24/1/26/1.1125,none/24/1/28/0.8,none/24/1/29/1.2,lightgray/33/1/35/0.8875,lightgray/33/1/35/1.1125}
\draw[black, fill=\style, thick] (\xa,\ya) rectangle (\xb,\yb);
\foreach \style/\xa/\ya/\xb/\yb in {none/33/1/39.25/1.2}
\draw[black, fill=\style, thick] (\xa,\ya) rectangle (\xb,\yb);
\foreach \y [evaluate = \y as \row using int((15-\y)*40)] in {1,...,3}
{
\draw (-0.25,\y) -- (39.25,\y);
\foreach \x in {0,...,39}
\draw[shift={(\x,\y)}] (0pt,2.5pt) -- (0pt,-2.5pt);
\foreach \x [evaluate = \x as \time using int(\x+\row-240)] in {0,5,...,35}
\draw (\x,\y) node [below=10] {\time};
\draw (0,\y) node [below=20] {Time};
}
\end{tikzpicture}
}
   \caption{Segment of a pacemaker trace showing blanking periods (gray bars), refractory periods (white bars), intrinsic beats (blue text), and paced events (green text). Atrial periods are shown above the timeline; ventricular are below. 
   }
  \label{fig_applicationTimelines}
\end{figure*}
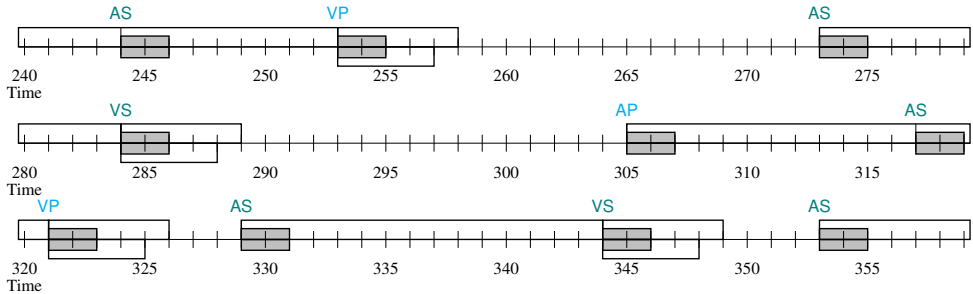

%% file: learning_rewards.tex

To enable a reinforcement learning (RL) agent to develop a control algorithm that satisfies a pacemaker's correctness requirements, we need a reward mechanism capable of classifying the actions predicted by the RL agent as correct or incorrect. Since pacing guidelines do not follow a deterministic process, the reward mechanism must be learned to ensure robustness. In this section, we propose a reward generator built using deep learning (DL). We begin by defining the learning task and then provide a description of the data used to train our reward machine.

\paragraph{Learning a Trace Classifier.}
We propose a rewards machine that learns to identify whether a pacemaker run was successful or not from its execution trace. A run will be considered successful if the pacemaker paces according to its requirements (see section \ref{pacemaker_requirements}), without missing or accidentally committing extra paces. 


There is a finite set of actions that a pacemaker can perform at a given clock tick: \textit{atrial pace}, \textit{ventricular pace}, and \textit{no action}. A pacemaker works by executing actions sequentially at each clock tick. Errors happen when the pacemaker commits an action extraneously, or when it misses an expected action. 
For example, an extraneous pace occurs if the expected action is a  \textit{no action} but the pacemaker performs a \textit{ventricular pace}---it has committed an action when it should have not done anything and is therefore erroneous.
Similarly, if the correct action to commit is an \textit{atrial pace} but the pacemaker performs a \textit{no action}, it has missed an action and committed an error through omission. Our main hypothesis is that a DL model that can learn to discriminate between positive and negative execution traces can be used to provide rewards to the RL agent. 

We construe the learning problem of the reward machine as a sequence classification task. Given a fixed context window of successful actions, we predict whether the next action is correct or not.  This problem can be formalized as: 

\begin{enumerate}
    \item Let $ \alpha_t \in \mathcal{A}$ be the action a pacemaker takes at clock tick $t$, from a finite set of actions $\mathcal{A}$.  
    \item Taking a trace with actions $\{ \alpha_0 , \alpha_1, \dots, \alpha_{t-1} \}$, the DL model predicts whether the next action $\alpha_t$ is correct (positive) or incorrect (negative). 
    \item We define $w$ as the sequence length/window size of the observed trace, and experiment with different sizes. 
\end{enumerate}







\paragraph{Pacemaker--Traces Dataset.}\label{pacemaker-traces-dataset}

In order to train deep neural networks for sequence classification on pacemaker traces, we construct a dataset using a pacemaker automaton and heart model. 
The automaton was generated from a DC specification~\cite{dole2023correct} for a dual chamber pacemaker and the heart model was created by the authors to exhibit healthy function or one of several common cardiac arrhythmias. 
The arrhythmias modeled were: sick sinus with complete AV block, sinus arrest, premature ventricular contractions (PVCs), Mobitz II (3:2 heart block), and stochastic. The pacemaker automaton was used as a surrogate for traces collected and annotated by a Clinician.
In addition to this, we also included pacing for a healthy heart. 
In each heart mode the intrinsic rate was allowed to stochastically drift up and down. 
Bad traces contained errors of omission, where a scheduled pacemaker generated pace should have occurred but did not, and extraneous, where additional pacemaker paces occurred in erroneous locations and the location and type of each error was randomly selected. 
See Figure \ref{fig_applicationTimelines} for a portion of a trace.

Following this process, we simulated a total of  $11,000$ samples.
Each arrhythmia type has $1,000$ negative samples (unsuccessful runs), and $1,000$ positive sample (successful runs). No good traces were generated for the AV heart block arrhythmia as all would have been exact copies of each other. This results in a total of $5,000$ positive examples, and $6,000$ negative examples (See table \ref{tab:dataset}).

\begin{table}[t!]
\caption{Positive and negative samples used for training rewards machine per arrhythmia type.}
\resizebox{\columnwidth}{!}{%
\begin{tabular}{@{}ccc@{}}
\toprule
\multicolumn{1}{c}{\textbf{Arrhythmia Type}} & \multicolumn{1}{c}{\textbf{\# Positive Samples}} & \textbf{\# Negative Samples} \\ \midrule
Complete AV block                             & -                                                 & $1000$                       \\
PVC                                           & $1000$                                            & $1000$                       \\
Mobitz II                                     & $1000$                                            & $1000$                       \\
Stochastic                                    & $1000$                                            & $1000$                       \\
Periodic Sinus Arrest                         & $1000$                                            & $1000$                       \\
Healthy Heart                                 & $1000$                                            & $1000$                       \\ \midrule
\textbf{Total}                                & \textbf{$5000$}                                   & \textbf{$6000$}              \\ \bottomrule
\end{tabular}%
}\label{tab:dataset}
\end{table}


While there are some cardiac waveform databases available~\cite{bhrs24,aaelectrogram24}, we were required to build our own datasets from scratch. Many available waveform databases are electrocardiogram (ECG) based. This is a body surface signal with a different morphology than the intracardiac electrogram (EGM) waveform measured by an implanted pacemaker. EGM libraries like~\cite{aaelectrogram24} contain analog waveforms measured at the pacemaker leads. Before such signals can be used by pacemaker pacing logic they are processed through a chain of analog and digital filters and comparators to remove noise sources such as muscle activity and breathing isolating the intrinsic pacing pulse. 
We do not attempt to model this functionality and assume this portion of the pacemaker has been properly configured to produce clean intrinsic activity.

%% file: experiments.tex
We present a cohesive case study showing how sequence learning and reinforcement learning can be combined into a unified workflow to learn a safety-critical task from trace examples, namely to operate as a pacemaker. The process entails two steps. First, learning a rewards machine from exemplary traces (box A of Figure \ref{fig:process}) followed by learning basic pacemaker functionality using the resultant rewards machine to critique the learning actions (box B of Figure \ref{fig:process}).

\subsection{Traces to Rewards}

In this subsection, we discuss our experimental study to test and evaluate our deep learning--based rewards machine. 
We first describe our dataset and model training paradigm, and then share our results.  

\subsubsection{Dataset and Experimental Setup.}
We distributed the positive and negative samples equally into $5$ folds. We used $3$ folds for training, and $1$ fold for validation, and $1$ fold of testing, which results in a total of $20$ different learning configurations. We report average and standard deviation metrics on the test for all configurations to assess model performance. We chose $4$ possible context sizes (i.e., the amount of prior observed steps in a trace), of $20$, $30$, $50$, $100$. We did not go any lower than $20$, as there may not be enough information in the sequence for the model to reliably learn the difference between correct and incorrect patterns. The rationale behind longer context sizes ($50$, $100$) is to explore if models could better use additional information, and if there is a trade--off due to the added variance that longer sequences bring.   

Thus, given a contextual sequence $\{ \alpha_1 , \alpha_2 \dots \alpha_{t-1} \}$ containing no pacing errors, the task is then to predict whether the next action $\alpha_t$ is correct or not.

\subsubsection{Model Training.}
We choose to test two different deep learning architectures that are commonly used for sequence classification: LSTMs and transformer networks. In order to ensure parity to both approaches, we try to maintain a similar set of hyper-parameters. For the LSTM, we implement a stacked bi-directional encoder model with 2 layers, where each hidden unit has a dimension of $256$. For the transformer, we implement 2 encoder--layers for attention. Each attention head has a hidden size of $2048$, and we use only one attention head per layer. 

We implement and train our models using PyTorch. We train each configuration for $10,000$ epochs with an ADAM optimizer, and a learning rate of $4.2e-6$. 
For each training epoch, we measure the model's performance on the validation set and save the model if there is an improvement, and report results of our best performing model. We use an NVIDIA A100 to train all models.


\subsubsection{Results.}
\begin{figure*}
     \centering
     \begin{subfigure}[b]{0.45\textwidth}
         \centering
         \includegraphics[width=0.8\textwidth]{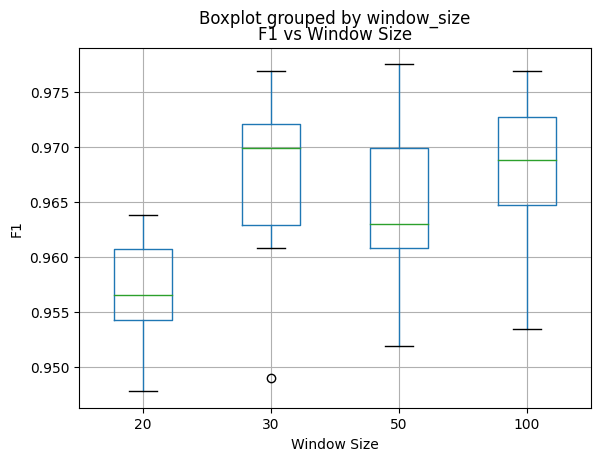}
         \caption{LSTM $F1$ scores across varying context sizes.}
         \label{fig:y equals x}
     \end{subfigure}
     \hfill
     \begin{subfigure}[b]{0.45\textwidth}
         \centering
         \includegraphics[width=0.8\textwidth]{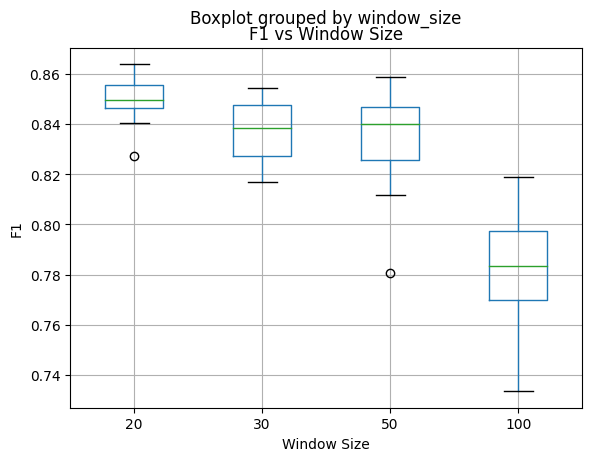}
         \caption{Transformer $F1$ scores across varying context sizes.}
         \label{fig:three sin x}
     \end{subfigure}
     \hfill
    \caption{Box plots, showing distribution of $F1$ scores across different context sizes for the $2$ different architectures.}
    \label{fig:boxplot-reward-machine}
\end{figure*}

The results from the training are given in Table \ref{tab:reward-machine-results}. 
The $P$ and $R$ columns stand for the precision and recall scores. 
We can see that the LSTM outperforms the transformer model significantly. 
The precision and recall scores are stable around the $F1$ scores, and predictions perform much better than chance ($0.5$, as we have two classes).   
These results can also be visualized in Figure \ref{fig:boxplot-reward-machine}.

\paragraph{Effects of context size on reward machine.}
The number of actions given as context to the deep learning model seems to have little effect in training the LSTM rewards generator. 
The $F1$ scores are quite comparable across all context lengths. However, when it comes to the transformer -- we observe that longer sequences perform poorly compared to shorter sequences. 

\paragraph{Effects of embedding layers on reward machine.}
It is common in deep learning to transform sparse representations to dense representations using an embedding layer. The embedding layer is a simple layer that transforms the sparse input (in our case, $1$--hot vectors of $5$--dimension, representing each of the actions that the pacemaker can take) into a dense vector. We ran a parallel experiment that utilizes an embedding layer, for half the number of epochs ($5000$) as our original experiment, which was enough for the model to reach convergence.  We did not find significant differences between using an embedding layer and otherwise. 

An explanation to why an embedding layer may not be beneficial to our learning problem could be because the input size is already quite small, and casting it to denser but larger representations may not capture additional information. 

\begin{table*}[t]
\caption{Reward Machine Training Results}
\label{tab:reward-machine-results}
\centering
\begin{tabular}{@{}c|lll|lll@{}}
\toprule[2.0pt]
\multirow{2}{*}{\textbf{Window Size}} & \multicolumn{3}{c|}{\textbf{LSTM}} & \multicolumn{3}{c}{\textbf{Transformer}}\\ 
 & \multicolumn{1}{c}{\textbf{P}}   & \multicolumn{1}{c}{\textbf{R}}   & \multicolumn{1}{c|}{\textbf{F1}}   & \multicolumn{1}{c}{\textbf{P}}   & \multicolumn{1}{c}{\textbf{R}}   & \multicolumn{1}{c}{\textbf{F1}}  \\ 
\midrule
20                   & 0.95 (0.003) & 0.95 (0.004) & 0.95 (0.004) & 0.86 (0.007) & 0.85 (0.007) & 0.85 (0.07) \\
30                   & 0.96 (0.007) & 0.96 (0.007) & 0.96 (0.007) & 0.84 (0.01)  & 0.83 (0.01)  & 0.83 (0.01) \\
50                   & 0.96 (0.007) & 0.96 (0.007) & 0.96 (0.007) & 0.83 (0.02)  & 0.83 (0.02)  & 0.83 (0.02) \\
100                  & 0.96 (0.005) & 0.96 (0.006) & 0.96 (0.006) & 0.80 (0.01)  & 0.78 (0.02)  & 0.78 (0.02) \\ 
\bottomrule[1.0pt]
\end{tabular}
\label{tbl:trainingResults}
\end{table*}

\begin{figure*}[t]
    \centering
    \input{ResultStateMachine}
    \caption{Learned State Machine - AS = Atrial Sense, VS = Ventricular Sense, AP = Atrial Pace, VP = Ventricular Pace}
    \label{fig:learnedDFA}
\end{figure*}

\begin{table}[
t]
\caption{Statistical Verification Results.}
\resizebox{\columnwidth}{!}{%
\begin{tabular}{@{}ccccc@{}}
\toprule
\multirow{2}{*}{\textbf{Arrhythmia Type}} & \textbf{\# Simulation} & \multicolumn{2}{c}{\textbf{Paces Generated}} & \textbf{Total} \\ 
& \textbf{Steps}& \textbf{AP} & \textbf{VP} & \textbf{Incorrect}\\
\midrule
Complete AV block                             & $50000$           & $1429$          & $1428$ & $0$\\
PVC                                           & $50000$           & $ 611$          & $ 643$ & $0$\\
Periodic Sinus Arrest                         & $50000$           & $1006$          & $ 475$ & $0$\\
Mobitz II                                     & $50000$           & $1345$          & $1106$ & $0$\\
Stochastic                                    & $50000$           & $ 675$          & $ 808$ & $0$\\
Healthy Heart                                 & $50000$           & $ 979$          &  $230$ & $0$\\ 
\midrule
\textbf{Total}                                & \textbf{$350000$}   & \textbf{$6045$} & \textbf{$4690$} & $0$\\ 
\bottomrule
\end{tabular}\label{tab:vvresults}%
}
\end{table}

\subsection{Rewards to Pacemaker}
We deploy a stochastic policy gradient algorithm with reward machines for learning pacemaker functionality. 
The reward machine does not work on individual actions but looks at a group of actions together and returns with a normalized reward based on only correctness of the final action in the group by first generating a log of pacemaker actions and intrinsic heart events at each step. When the log is full, it is graded by the rewards agent and the log is added to the replay memory. If the rewards agent returned a low reward meaning actions in the log were incorrect, the MDP is reset and the simulation restarted. When the replay memory is full, the episode is over and the pacemaker agent is trained using the replay memory. The replay memory is then cleared and a new episode begins.

There are multiple parameters that control how the pacemaker agent learns. The exploration rate ($\gamma$) specifies how often a random action is selected. The log length defines how many steps are taken before grading the selected actions. The number of logs to be combined before training is defined by the replay memory size and episodes specifies the duration of training.

Training was performed a total of eight times, four each for each type of reward machine (LSTM and transformer) varying the window sizes used as shown in Table~\ref{tbl:trainingResults}. For each pacing agent training, the reward machines were trained with 20 folds and the pacing agent's log depth was set equal to the reward machines window size. The replay memory consisted of 200 logs and training ran for 50,000 episodes. The heart model was set to stochastic mode to assure the pacemaker agent was exposed to the widest possible variations in heart actions.

Validation was done using two methods. The first used the trained agent in prediction mode with the heart in each disease state. Table~\ref{tab:vvresults} shows the results of the testing. The tables shows the number of simulation steps taken along with the number of atrial and ventricular paces generated and the total incorrectly generated paces. Over 350,000 steps, representing approximately $2.75$ continuous hours of real-time operation, no erroneous or omitted paces were observed. 
The second validation was through extraction of the learned controller. 
Seen in Figure~\ref{fig:learnedDFA}, each agent returned the same state machine and these were inspected by a pacemaker designer for accuracy. 
In the shown diagram, each state name is an encoding of the currently active periods, intervals, and intrinsic heart beats as described in Table~\ref{tbl:StateNames}. The graph edges represent the pacemaker agent's action (AP = Atrial Pace, VP = Ventricular Pace, -- = Wait) and there is only one valid action that can be taken in each state. There are two state transitions that may look incorrect. State 11000000, where a VS had occurred, is exited by an atrial pace. Likewise, state 00100000 exits with a VP even though an AS had occurred. This is acceptable behavior for a pacemaker. When an AV or VA interval ends, the pacemaker will pace regardless of heart activity. Doing otherwise would delay the required pacing pulse and violate the specification by pacing the patient below LRI. A check could be added before pace delivery but the intrinsic event could still fall between the check and the pace; there's no practical way to fully mitigate this.

\begin{table}
\caption{Decoding Pacemaker State Name}
\centering
\begin{tabular}{c|l}
    \toprule
    Bit & Description \\
    \midrule
    b7 & Interval \\
    b6 & Ventricular Sense \\
    b5 & Atrial Sense \\
    b4 & AV or VA Interval Running \\
    b3 & Atrial Refractory (AR) Period \\
    b2 & Atrial Blanking (AB) Period \\
    b1 & Ventricular Refractory (VR) Period \\
    b0 & Ventricular Blanking (VB) Period \\
    \bottomrule
\end{tabular}\label{tbl:StateNames}

\end{table}

%% file: ResultStateMachine.tex
\adjustbox{max width=12                                                                                                                                                                                          cm}{
\begin{tikzpicture}[>=latex',line join=bevel, line width=2pt, font=\sffamily]
\begin{scope}
  \pgfsetstrokecolor{black}
  \definecolor{strokecol}{rgb}{0.66,0.66,0.66};
  \pgfsetstrokecolor{strokecol}
  \draw [line width=3pt] (8.0bp,105.0bp) -- (8.0bp,542.0bp) -- (702.6bp,542.0bp) -- (702.6bp,105.0bp) -- cycle;
  \definecolor{strokecol}{rgb}{0.0,0.0,0.0};
  \pgfsetstrokecolor{strokecol}
  \draw (355.3bp,529.75bp) node {VA Interval};
\end{scope}
\begin{scope}
  \pgfsetstrokecolor{black}
  \definecolor{strokecol}{rgb}{0.66,0.66,0.66};
  \pgfsetstrokecolor{strokecol}
  \draw [line width=3pt] (740.6bp,42.0bp) -- (740.6bp,446.0bp) -- (1150.8bp,446.0bp) -- (1150.8bp,42.0bp) -- cycle;
  \definecolor{strokecol}{rgb}{0.0,0.0,0.0};
  \pgfsetstrokecolor{strokecol}
  \draw (945.7bp,433.75bp) node {AV Interval};
\end{scope}
  \node (10011111) at (48.4bp,165.0bp) [draw=black,circle, align=center] {10011111};
  \node (10011010) at (158.2bp,260.0bp) [draw=black,circle, align=center] {10011010};
  \node (11011010) at (268.0bp,312.0bp) [draw=black,circle, align=center] {11011010\\ \textcolor{black!20!green}{VS}};
  \node (10111010) at (268.0bp,179.0bp) [draw=black,circle, align=center] {10111010\\ \textcolor{black!20!green}{VS}};
  \node (10011000) at (433.6bp,238.0bp) [draw=black,circle, align=center] {10011000};
  \node (11011000) at (268.0bp,436.0bp) [draw=black,circle, align=center] {11011000\\ \textcolor{black!20!green}{VS}};
  \node (10111000) at (350.8bp,353.0bp) [draw=black,circle, align=center] {10111000\\ \textcolor{black!20!green}{AS}};
  \node (10010000) at (543.4bp,328.0bp) [draw=black,circle, align=center] {10010000};
  \node (10110000) at (662.2bp,228.0bp) [draw=black,circle, align=center] {10110000\\ \textcolor{black!20!green}{AS}};
  \node (11010000) at (662.2bp,145.0bp) [draw=black,circle, align=center] {11010000\\ \textcolor{black!20!green}{VS}};
  \node (10000000) at (662.2bp,394.0bp) [draw=black,circle, align=center] {10000000};
  \node (11000000) at (662.2bp,311.0bp) [draw=black,circle, align=center] {11000000\\ \textcolor{black!20!green}{VS}};
  \node (10100000) at (662.2bp,477.0bp) [draw=black,circle, align=center] {10100000\\ \textcolor{black!20!green}{AS}};
  \node (00011101) at (781.0bp,277.0bp) [draw=black,circle, align=center] {00011101};
  \node (00011000) at (890.8bp,277.0bp) [draw=black,circle, align=center] {00011000};
  \node (01011000) at (1000.6bp,165.0bp) [draw=black,circle, align=center] {01011000\\ \textcolor{black!20!green}{VS}};
  \node (00111000) at (1000.6bp,248.0bp) [draw=black,circle, align=center] {00111000\\ \textcolor{black!20!green}{AS}};
  \node (00000000) at (1110.4bp,248.0bp) [draw=black,circle, align=center] {00000000};
  \node (01000000) at (1000.6bp,82.0bp) [draw=black,circle, align=center] {01000000\\ \textcolor{black!20!green}{VS}};
  \node (00100000) at (1000.6bp,381.0bp) [draw=black,circle, align=center] {00100000\\ \textcolor{black!20!green}{AS}};
  \draw [black,->] (10011111) ..controls (38.0bp,206.74bp) and (41.43bp,215.4bp)  .. (48.4bp,215.4bp) .. controls (52.647bp,215.4bp) and (55.58bp,212.18bp)  .. (10011111);
  \definecolor{strokecol}{rgb}{0.0,0.0,1.0};
  \pgfsetstrokecolor{strokecol}
  \draw (48.4bp,222.9bp) node {\textcolor{black!10!blue}{--}};
  \draw [black,->] (10011111) ..controls (88.537bp,199.49bp) and (108.42bp,217.01bp)  .. (10011010);
  \draw (103.3bp,222.5bp) node {\textcolor{black!10!blue}{--}};
  \draw [black,->] (10011111) ..controls (69.506bp,232.3bp) and (91.777bp,291.94bp)  .. (125.8bp,334.0bp) .. controls (154.55bp,369.54bp) and (165.18bp,380.73bp)  .. (208.6bp,395.0bp) .. controls (231.49bp,402.52bp) and (246.59bp,378.32bp)  .. (11011010);
  \draw (158.2bp,395.5bp) node {\textcolor{black!10!blue}{--}};
  \draw [black,->] (10011111) ..controls (118.75bp,169.46bp) and (181.94bp,173.53bp)  .. (10111010);
  \draw (158.2bp,181.5bp) node {\textcolor{black!10!blue}{--}};
  \draw [black,->] (10011010) ..controls (147.8bp,301.74bp) and (151.23bp,310.4bp)  .. (158.2bp,310.4bp) .. controls (162.45bp,310.4bp) and (165.38bp,307.18bp)  .. (10011010);
  \draw (158.2bp,317.9bp) node {\textcolor{black!10!blue}{--}};
  \draw [black,->] (10011010) ..controls (239.91bp,253.51bp) and (334.49bp,245.89bp)  .. (10011000);
  \draw (268.0bp,260.5bp) node {\textcolor{black!10!blue}{--}};
  \draw [black,->] (10011010) ..controls (199.74bp,262.9bp) and (209.32bp,265.16bp)  .. (217.6bp,269.0bp) .. controls (224.39bp,272.15bp) and (230.96bp,276.53bp)  .. (11011010);
  \draw (213.1bp,276.5bp) node {\textcolor{black!10!blue}{--}};
  \draw [black,->] (10011010) ..controls (180.3bp,311.44bp) and (193.68bp,340.33bp)  .. (208.6bp,364.0bp) .. controls (217.22bp,377.67bp) and (228.06bp,391.69bp)  .. (11011000);
  \draw (213.1bp,383.5bp) node {\textcolor{black!10!blue}{--}};
  \draw [black,->] (10011010) ..controls (199.11bp,230.01bp) and (217.23bp,216.39bp)  .. (10111010);
  \draw (213.1bp,228.5bp) node {\textcolor{black!10!blue}{--}};
  \draw [black,->] (10011010) ..controls (186.47bp,310.77bp) and (207.64bp,339.78bp)  .. (235.6bp,353.0bp) .. controls (257.73bp,363.46bp) and (285.08bp,363.83bp)  .. (10111000);
  \draw (213.1bp,347.5bp) node {\textcolor{black!10!blue}{--}};
  \draw [black,->] (10011000) ..controls (473.95bp,270.85bp) and (493.15bp,286.89bp)  .. (10010000);
  \draw (488.5bp,292.5bp) node {\textcolor{black!10!blue}{--}};
  \draw [black,->] (10011000) ..controls (505.74bp,234.86bp) and (573.95bp,231.85bp)  .. (10110000);
  \draw (543.4bp,241.5bp) node {\textcolor{black!10!blue}{--}};
  \draw [black,->] (10011000) ..controls (503.85bp,209.58bp) and (576.12bp,179.92bp)  .. (11010000);
  \draw (543.4bp,212.5bp) node {\textcolor{black!10!blue}{--}};
  \draw [black,->] (11011010) ..controls (227.98bp,296.85bp) and (217.77bp,292.46bp)  .. (208.6bp,288.0bp) .. controls (204.62bp,286.07bp) and (200.53bp,283.95bp)  .. (10011010);
  \draw (213.1bp,298.5bp) node {\textcolor{black!10!blue}{--}};
  \draw [black,->] (11011010) ..controls (324.34bp,286.97bp) and (363.77bp,269.13bp)  .. (10011000);
  \draw (350.8bp,295.5bp) node {\textcolor{black!10!blue}{--}};
  \draw [black,->] (11011000) ..controls (203.86bp,438.5bp) and (155.72bp,433.98bp)  .. (125.8bp,407.0bp) .. controls (69.254bp,356.01bp) and (54.236bp,263.58bp)  .. (10011111);
  \draw (158.2bp,441.5bp) node {\textcolor{black!10!blue}{--}};
  \draw [black,->] (10010000) ..controls (533.0bp,369.74bp) and (536.43bp,378.4bp)  .. (543.4bp,378.4bp) .. controls (547.65bp,378.4bp) and (550.58bp,375.18bp)  .. (10010000);
  \draw (543.4bp,385.9bp) node {\textcolor{black!10!blue}{--}};
  \draw [black,->] (10010000) ..controls (585.91bp,292.46bp) and (609.4bp,272.35bp)  .. (10110000);
  \draw (602.8bp,290.5bp) node {\textcolor{black!10!blue}{--}};
  \draw [black,->] (10010000) ..controls (576.9bp,271.12bp) and (604.04bp,225.2bp)  .. (629.8bp,187.0bp) .. controls (631.4bp,184.62bp) and (633.09bp,182.19bp)  .. (11010000);
  \draw (602.8bp,247.5bp) node {\textcolor{black!10!blue}{--}};
  \draw [black,->] (10010000) ..controls (587.64bp,352.44bp) and (607.28bp,363.53bp)  .. (10000000);
  \draw (602.8bp,371.5bp) node {\textcolor{black!10!blue}{--}};
  \draw [black,->] (10010000) ..controls (588.92bp,321.52bp) and (604.53bp,319.25bp)  .. (11000000);
  \draw (602.8bp,327.5bp) node {\textcolor{black!10!blue}{--}};
  \draw [black,->] (10010000) ..controls (583.51bp,377.89bp) and (612.98bp,415.49bp)  .. (10100000);
  \draw (602.8bp,418.5bp) node {\textcolor{black!10!blue}{--}};
  \draw [black,->] (10111010) ..controls (226.68bp,188.3bp) and (216.75bp,192.35bp)  .. (208.6bp,198.0bp) .. controls (198.85bp,204.76bp) and (190.06bp,214.01bp)  .. (10011010);
  \draw (213.1bp,205.5bp) node {\textcolor{black!10!blue}{--}};
  \draw [black,->] (10111010) ..controls (325.18bp,199.26bp) and (363.38bp,213.04bp)  .. (10011000);
  \draw (350.8bp,225.5bp) node {\textcolor{black!10!blue}{--}};
  \draw [black,->] (11010000) ..controls (558.98bp,137.57bp) and (382.65bp,127.39bp)  .. (235.6bp,138.0bp) .. controls (185.85bp,141.59bp) and (129.2bp,150.5bp)  .. (10011111);
  \draw (350.8bp,141.5bp) node {\textcolor{black!10!blue}{--}};
  \draw [black,->] (10111000) ..controls (415.0bp,344.71bp) and (464.2bp,338.26bp)  .. (10010000);
  \draw (433.6bp,353.5bp) node {\textcolor{black!10!blue}{--}};
  \draw [black,->] (00011101) ..controls (824.18bp,277.0bp) and (835.93bp,277.0bp)  .. (00011000);
  \draw (835.9bp,284.5bp) node {\textcolor{black!10!blue}{--}};
  \draw [black,->] (00011101) ..controls (813.02bp,230.21bp) and (834.16bp,203.84bp)  .. (858.4bp,187.0bp) .. controls (893.65bp,162.51bp) and (907.41bp,153.56bp)  .. (950.2bp,157.0bp) .. controls (952.41bp,157.18bp) and (954.67bp,157.4bp)  .. (01011000);
  \draw (890.8bp,194.5bp) node {\textcolor{black!10!blue}{--}};
  \draw [black,->] (00011101) ..controls (819.33bp,242.72bp) and (838.58bp,228.56bp)  .. (858.4bp,221.0bp) .. controls (896.67bp,206.4bp) and (911.18bp,200.58bp)  .. (950.2bp,213.0bp) .. controls (955.75bp,214.77bp) and (961.22bp,217.41bp)  .. (00111000);
  \draw (890.8bp,228.5bp) node {\textcolor{black!10!blue}{--}};
  \draw [black,->] (00011000) ..controls (880.4bp,318.74bp) and (883.83bp,327.4bp)  .. (890.8bp,327.4bp) .. controls (895.05bp,327.4bp) and (897.98bp,324.18bp)  .. (00011000);
  \draw (890.8bp,334.9bp) node {\textcolor{black!10!blue}{--}};
  \draw [black,->] (00011000) ..controls (930.01bp,237.3bp) and (952.06bp,214.39bp)  .. (01011000);
  \draw (945.7bp,231.5bp) node {\textcolor{black!10!blue}{--}};
  \draw [black,->] (00011000) ..controls (927.89bp,290.23bp) and (934.7bp,291.97bp)  .. (941.2bp,293.0bp) .. controls (965.78bp,296.88bp) and (923.95bp,319.23bp)  .. (1033.0bp,289.0bp) .. controls (1046.8bp,285.17bp) and (1060.9bp,278.51bp)  .. (00000000);
  \draw (1000.6bp,311.5bp) node {\textcolor{black!10!blue}{--}};
  \draw [black,->] (00011000) ..controls (932.2bp,276.56bp) and (941.68bp,275.35bp)  .. (950.2bp,273.0bp) .. controls (954.02bp,271.95bp) and (957.9bp,270.59bp)  .. (00111000);
  \draw (945.7bp,281.5bp) node {\textcolor{black!10!blue}{--}};
  \draw [black,->] (00011000) ..controls (928.27bp,203.88bp) and (967.75bp,124.73bp)  .. (968.2bp,124.0bp) .. controls (969.71bp,121.56bp) and (971.32bp,119.08bp)  .. (01000000);
  \draw (945.7bp,180.5bp) node {\textcolor{black!10!blue}{--}};
  \draw [black,->] (00011000) ..controls (930.56bp,314.39bp) and (951.38bp,334.48bp)  .. (00100000);
  \draw (945.7bp,339.5bp) node {\textcolor{black!10!blue}{--}};
  \draw [black,->] (00111000) ..controls (959.23bp,250.19bp) and (949.75bp,251.64bp)  .. (941.2bp,254.0bp) .. controls (937.64bp,254.98bp) and (934.02bp,256.2bp)  .. (00011000);
  \draw (945.7bp,261.5bp) node {\textcolor{black!10!blue}{--}};
  \draw [black,->] (00111000) ..controls (1043.8bp,248.0bp) and (1055.5bp,248.0bp)  .. (00000000);
  \draw (1055.5bp,255.5bp) node {\textcolor{black!10!blue}{--}};
  \draw [black,->] (10110000) ..controls (707.06bp,246.4bp) and (724.58bp,253.75bp)  .. (00011101);
  \draw (721.6bp,262.5bp) node {\textcolor{black!10!blue}{--}};
  \draw [black,->] (10000000) ..controls (703.9bp,353.23bp) and (729.65bp,327.44bp)  .. (00011101);
  \draw (721.6bp,349.5bp) node {\textcolor{black!10!blue}{AP}};
  \draw [black,->] (01011000) ..controls (961.47bp,115.86bp) and (928.86bp,85.0bp)  .. (891.8bp,85.0bp) .. controls (720.6bp,85.0bp) and (720.6bp,85.0bp)  .. (720.6bp,85.0bp) .. controls (617.75bp,85.0bp) and (592.35bp,97.0bp)  .. (489.5bp,97.0bp) .. controls (157.2bp,97.0bp) and (157.2bp,97.0bp)  .. (157.2bp,97.0bp) .. controls (127.52bp,97.0bp) and (99.341bp,115.67bp)  .. (10011111);
  \draw (543.4bp,103.5bp) node {\textcolor{black!10!blue}{--}};
  \draw [black,->] (11000000) ..controls (707.26bp,298.17bp) and (723.63bp,293.41bp)  .. (00011101);
  \draw (721.6bp,303.5bp) node {\textcolor{black!10!blue}{AP}};
  \draw [black,->] (00000000) ..controls (1108.2bp,147.98bp) and (1094.4bp,0.0bp)  .. (1001.6bp,0.0bp) .. controls (157.2bp,0.0bp) and (157.2bp,0.0bp)  .. (157.2bp,0.0bp) .. controls (98.789bp,0.0bp) and (70.345bp,73.365bp)  .. (10011111);
  \draw (602.8bp,7.5bp) node {\textcolor{black!10!blue}{VP}};
  \draw [black,->] (01000000) ..controls (950.35bp,57.233bp) and (920.02bp,46.0bp)  .. (891.8bp,46.0bp) .. controls (157.2bp,46.0bp) and (157.2bp,46.0bp)  .. (157.2bp,46.0bp) .. controls (113.75bp,46.0bp) and (83.245bp,90.118bp)  .. (10011111);
  \draw (543.4bp,53.5bp) node {\textcolor{black!10!blue}{VP}};
  \draw [black,->] (10100000) ..controls (691.94bp,448.81bp) and (697.83bp,441.91bp)  .. (702.6bp,435.0bp) .. controls (728.26bp,397.86bp) and (750.09bp,350.94bp)  .. (00011101);
  \draw (721.6bp,426.5bp) node {\textcolor{black!10!blue}{AP}};
  \draw [black,->] (00100000) ..controls (980.92bp,458.15bp) and (951.72bp,537.0bp)  .. (891.8bp,537.0bp) .. controls (157.2bp,537.0bp) and (157.2bp,537.0bp)  .. (157.2bp,537.0bp) .. controls (89.598bp,537.0bp) and (61.687bp,307.5bp)  .. (10011111);
  \draw (543.4bp,544.5bp) node {\textcolor{black!10!blue}{VP}};
\end{tikzpicture}
}

%% file: related.tex
In 2007, in conjunction with the release of~\citet{paceSpec2009}, the North American Software Certification Consortium (SCC) initiated the Pacemaker Formal Methods Challenge
hosted by the McMaster University Software Quality Research Lab (SQRL). As a progress report on the challenge, Schloss Dagstuhl convened a seminar in 2012 to review the results of the challenge to date~\cite{mery2014pacemaker}. A review of formal methods usage in medical devices was published in 2018~\cite{bonfanti2018systematic}. Table 4 of that document provides a breakdown of 48 papers related to pacemaker research into categories of Modeling, Model Verification, Model Validation, Software Validation, and Code Generation. Specific to our work, there are several works related to translating pacemaker requirements into a formal logic. Unlike our approach of extracting expected functionality from examples, each of these requires the human designer to perform the translation from natural language to formal logic.

\citet{mery2011formal} formalizes \cite{paceSpec2009} in the EVENT B modeling language, which provided structure for arbitrary clocks, freeing the authors from the real-time limitation of predicate logic and found during verification that static proof analysis was not sufficient to assure proper operation and functional modeling was required to identify deadlocks and logic failures.

\citet{gomes2009formal} developed a formal pacemaker specification in the Z language~\cite{lightfoot1991formal}, a specification language built on formal proofs of propositional and predicate logic with no inherent construct for representing hard real-time limits and arbitrary, stochastic time periods. As presented in their paper, they were able to specify and verify a simple dual chamber (DDD \cite{bernstein2002revised}) pacing system. Their specification was, however, incomplete as it lacked fundamental aspects of safe pacing, namely refractory and blanking timing and rate response. 

\citet{larson2014formal} created a version of the pacemaker specification using the BLESS~\cite{larson2013bless} behavioral language, an extension of AADL. With the BLESS syntax, a complete model was able to be created for the entire specification. Verification of satisfaction to the original specification was never proven. 

\citet{jiang2012modeling} developed a pacemaker model using UPPAAL~\cite{bengtsson1996uppaal}. Their work provided extensive verification of complex pacemaker features, including pacemaker-mediated tachycardia (PMT) and mode switching. The UPPAAL model was created manually.
\citet{dole2023correct} used Duration Calculus (DC) logic to express pacemaker specifications in formal real-time logic, which were then converted into finite state machines (automata). These automata were subsequently used as the reward machine in training a reinforcement learning (RL) agent. Validation of the agent's functionality demonstrated a viable, complete process starting from a formal language through to implementation and validation. However, their work did not address errors that could occur during the translation from specification to logic. Moreover, \citet{dole2023correct} highlighted the difficulty of translating requirements from natural language to formal logic.
In this paper, our goal is to demonstrate a framework that can directly learn pacemaker logic from expert demonstrations, obviating the need for logic experts to translate requirements into formal logic.

%% file: conclusions.tex
Our results show that it is possible to both train an RL agent to recognize correct operation of a complex medical device solely from examples and use that agent in training another agent to properly operate that medical device. We did not directly address safety guarantees with this work and point work on Shielded RL~\cite{dole2023correct,alshiekh2018safe} and formal verification~\cite{jiang2012modeling} for well-established solutions. 
While current RL methods cannot guarantee safe operation in a medical device, this work does point the way to a new design paradigm where subject matter experts (SMEs) are directly designing the product through creation of examples rather than dictating requirements to a designer. Designing directly from examples removes the knowledge gap between the designer and SME where requirements hand-off leads to misinterpretation.


While the research we presented here focused on a pacemaker as a real world example of complex design, the field of medical device, both implanted and external, is vast. Examples of such devices cover insulin and drug pump, neurological stimulation devices for pain and tremor reduction, mobility devices such as robotics and prosthetics, and intelligent surgical devices. The design challenges to translate clinician and medical research knowledge into devices that provide safe and effective treatment are similar to that of a pacemaker.

%% file: appendix.tex
\section{Supplemental Information}
\subsection{Duration Calculus}
Let $\Var$ be a finite set of atomic propositions.
We interpret DC formulas over traces generated from runs of systems.
Given the set of variables $\Var$, we define the syntax of DC formulas (following~\cite{dole2023correct}) as follows:
\begin{eqnarray*}
  P &::=& \false \mid \true \mid x \in \Var \mid P \wedge P \mid \neg P\\
	D &::=& \rangeP{P}   \mid \pointP{P} \mid  D \land D \mid \neg D \mid
		  D \chop D \mid M\\
	M & ::= & \ell \mop c \mid \int P \mop c \mid  \sum P \mop c 
\end{eqnarray*}

DC formulas are evaluated over timed traces $\sigma$ and a reference  interval $I = [b, e]$ where $b \leq e$ and $b, e \in \mathbb{N}$ range over the indices $0, \dots, n$ of the timed trace $\sigma=(s_0, \tau_0) \dots (s_n, \tau_n)$. 
The satisfaction of a DC formula $\psi$ evaluated on timed trace  $\sigma = \seq{(s_0, \tau_0), (s_1, \tau_1), \ldots}$  with respect to an interval $[b,e]$ and is denoted as $(\sigma, [b,e]) \models \psi$.

For a timed trace $\sigma=\seq{(s_0,\tau_0) (s_1, \tau_1) \dots (s_n, \tau_n)}$ and propositional formula $P$, we say $(\sigma, i) \models P$ iff $s_i \models P$. 
The satisfaction of other DC formulas is defined inductively:
  \begin{enumerate}
  \item
    $(\sigma, [b,e]) \models  \rangeP{P}$ iff $b{<}e$, and 	$(\sigma, t) \models P$ for all  $b {<} t {<} e$;
    \item  $(\sigma, [b,e]) \models  \pointP{P}$ iff $b{=}e$ and  $(\sigma, b) \models P$;
  \item $(\sigma, [b,e]) \models  D_1 {\wedge} D_2 $ iff $(\sigma, [b,e]) {\models} D_1$,$(\sigma, [b,e]) {\models} D_2$;
  \item  $(\sigma, [b,e]) \models \neg D$ iff $(\sigma, [b,e]) \not \models D$;
  \item $(\sigma, [b,e]) \models  D_1 \chop D_2$ iff there exists a point $b \leq z \leq e$ s.t. $(\sigma,[b,z]) \models D_1$ and  $(\sigma,[z,e]) \models D_2$;
  \item $(\sigma, [b,e]) \models$ $\ell \mop c$ iff $(\tau_e-\tau_b) \mop c$ holds;
  \item $(\sigma, [b,e]) \models {\int} P {\mop} c$ iff 
  $\sum \set{\tau_{i+1} {-} \tau_i \::\: (\sigma, i) \models P} \mop c$;
\item $(\sigma, [b,e]) \models \sum P \mop c$ iff 
$|\set{i \::\: (\sigma, i) \models P}| \mop c$;

\end{enumerate}
For decidability, DDC eliminates the $\int P$ and $\ell$ operators. 
Using the chop modality $\chop$, one can derive the following syntactic sugar:
eventually modality $\Diamond D \rmdef \true \chop D \chop \true$ and 
  globally modality $\square  D \rmdef \neg \Diamond  \neg D$.
 One can also define the integral duration modality  $\ell \in \Nat$ to denote an  integral interval, that is, 
 $(\sigma, [b,e]) \models \ell \in \Nat$ iff $\tau_e-\tau_b \in \Nat$. 
 

\subsubsection{Bradycardia Pacing Requirements}
An example of a natural language pacemaker requirement and its DC equivalent is as follows:
\begin{description}
     \item \emph{The lower rate limit interval (LRL) and the upper rate limit interval (URL) start at a paced ventricular event or non-refractory ventricular sensed event.}
\end{description}

Translates to the DC formula:
\begin{equation}
\begin{split}
 (\ftrigInt{(VP \lor VSN)}{\llimit}{LRL}) \land\\
 (\ftrigInt{(VP \lor VSN)}{\ulimit}{URL}) \land\\
  (\ftrigInt{(AP \lor ASN)}{\llimit}{ALRL}) \land\\
 (\ftrigInt{(AP \lor ASN)}{\ulimit}{AURL})
\end{split}
\end{equation}
Where a ventricular pace VP or sense VSN triggers the start of an LRL interval (\textit{Represented as: }$\ftrigInt{}{\llimit}{LRL}$) and an URL interval while an atrial pace (AP) or sense (ASN) triggers an atrial initiated LRL interval and URL interval. 
We refer to the appendix of \cite{dole2023correct} for a full list of the DC pacemaker specifications.

In the absence of a definite source of labeled pacemaker traces, we use automata generated from the work of Dole et al. \cite{dole2023correct} to extract expert demonstrations to train our reward machine.

\subsection{Reinforcement Learning}
Reinforcement learning (RL)\cite{sutton2018reinforcement} is a sampling-based algorithm to sequential optimization where an agent seeks learn an optimal policy guided a reward signal. 
Recent RL algorithms use function approximators
(deep neural networks) and build upon the \emph{policy gradient theorem}
~\cite{sutton_policy_2000} to directly optimize the parameterized
controllers~\cite{silver_deterministic_2014,haarnoja_soft_2018}.
The results of the theorem allows for the computation of an approximate
gradient over the parameter space of the neural network, without having to
explicitly compute the expected rewards or \(Q-\)function by enumerating the
state-action pairs.

\begin{algorithm}[ht!]
\caption{Pacemaker Agent Learning From Trace Log}
\label{alg:Pacemaker}
\textbf{Input}: Episode Cnt ($E$), Replay Size($R$), Log Len ($L$), Explore Rate ($\gamma$)  \\
\textbf{Output}: Trained RL Agent
    \begin{algorithmic}[1] 
        \Procedure{fit}{$E, L, \gamma$}
        \For {each Episode}
            \State $ State \gets S_0$
            \State $Replay \gets NewList()$
            \For {each Replay}
                \State $log \gets NewList()$.
                \For {each step in Log}
                    \State $Action \gets NextAction(State,\gamma)$
                    \State $State, Event \gets Step(Action)$
                    \State $log.AddItem([Action,Event])$
                \EndFor
                \State $Done \gets Grade(log)$
                \State $Replay.AddItem(log, Done)$
                \If{$Done$}
                    \State $State \gets S_0$
                \EndIf
            \EndFor
            \State $Train(Replay)$
        \EndFor
        \EndProcedure
    \end{algorithmic}
\end{algorithm}

Let $D(A)$ be the set of distributions over a set $A$.
A \emph{Markov decision process} $M$ is a tuple $(S, A, T)$ where $S$ is a set of states, $A$ is a finite set of {\it actions}, and $T : S \times A \to
	D(S)$ is the {probabilistic transition function}.
For any state $s \in S$, we let $A(s)$ denote the set of actions that can be selected in state $s$.  
For states $s, s' \in S$ and $a \in A(s)$, $T(s,
	a)(s')$ equals $p(s' | s, a)$.  
 A \emph{run} of $M$ is an infinite sequence 
$(s_0, a_1, s_1, \ldots)$ such that
$p(s_{i+1} | s_{i}, a_{i+1}) {>} 0$ for all $i \geq 0$. 
A finite run is a finite such sequence. 
We write $Runs^M (FRuns^M)$  for the set of
runs (finite runs) of the MDP $M$.

A \emph{reward machine} implements a function $r: Runs \to \mathbb{R}$ that provides a scalar reward for every finite run of the MDP. Typically, reward machines are implemented via finite state machines~\cite{icarte2022reward}. In this work, we learn a reward machine from expert demonstrations as a deep neural network.

Let \(\pi_\theta: S \to D(A)\) denote a stationary strategy for the MDP $M$ that is parameterized by \(\pi\). We define our
reinforcement learning problem as solving an optimization problem that
maximizes the expected long-term reward received by the strategy
\(\pi_\theta\) on the augmented MDP $M$. We are 
interested in maximizing the finite horizon ($T < \infty$) objective 
$J(\theta) = \mathbb{E}_{\pi_\theta} \left[ G_{0,T} \right]$
where $G_{0,T} = \sum_{t = 0}^T \lambda^{t} r_t$ and \(\lambda \in [0,1)\) is some discount factor and \(r_t\) is the
reward obtained at time step \(t\) by taking some action \(a_t\) from the
strategy \(\pi_\theta\) at state \(s_t\). 
Policy Gradient seeks to refine a policy via gradient ascent given a set of experience acquired from a given policy. 
The goal is to find the optimal policy by adjusting $\theta$ to maximize the objective function $J(\theta$) using gradient ascent for a gradient step size $\alpha$.
\begin{align}
    \theta_{new}\longleftarrow \theta + \alpha \nabla_\theta J(\theta)
\end{align}
The gradient of the objective function is the expected return of the discounted rewards $G_0$ starting from the initial state $s_0$ over a trace $\tau$ of length $T$ per policy $\pi_\theta$:
\begin{align}
    \nabla_\theta J(\theta) = \mathbb E_\tau \left [ G_{0,T} \sum_{t=0}^T \nabla_\theta \log \pi_\theta(a_t | s_t)\right ].
\end{align}

Algorithm~\ref{alg:Pacemaker} shows the basic structure of the policy gradient algorithm used to learn the functionality of the pacemaker for the case study in this paper.